\documentclass{bmvc2k}

\usepackage{amsmath}
\usepackage{amssymb}
\usepackage{bigstrut}
\usepackage{multirow}
\usepackage{booktabs}
\usepackage[misc]{ifsym}

\title{Distribution-Aware Calibration for Object Detection with Noisy Bounding Boxes}

\addauthor{Donghao Zhou}{dhzhou@link.cuhk.edu.hk}{1}
\addauthor{Jialin Li}{lijialinthu@gmail.com}{2}
\addauthor{Jinpeng Li}{1155074899@link.cuhk.edu.hk}{1}
\addauthor{Jiancheng Huang}{jianchenghuang@smail.nju.edu.cn}{3}
\addauthor{Qiang Nie}{qiangnie@hkust-gz.edu.cn}{4,2}
\addauthor{Yong Liu}{choasliu@tencent.com}{2 \Letter}
\addauthor{Bin-Bin Gao}{csgaobb@gmail.com}{2}
\addauthor{Qiong Wang}{wangqiong@siat.ac.cn}{5}
\addauthor{Pheng-Ann Heng}{pheng@cse.cuhk.edu.hk}{1}
\addauthor{Guangyong Chen}{gychen@zhejianglab.com}{6 \Letter}

\addinstitution{
 The Chinese University of Hong Kong \\
 Hong Kong, China
}
\addinstitution{
 Tencent YouTu Lab \\
 Shenzhen, China
}
\addinstitution{
 Shenzhen Institute of Advanced Technology, Chinese Academy of Sciences \\
 Shenzhen, China
}
\addinstitution{
 The Hong Kong University of Science and Technology (Guangzhou) \\
 Guangzhou, China
}
\addinstitution{
Guangdong Provincial Key Laboratory of Computer Vision and Virtual Reality Technology, Shenzhen Institute of Advanced Technology, Chinese Academy of Sciences \\
 Shenzhen, China
}
\addinstitution{
 Zhejiang Lab \\
 Hangzhou, China
}

\runninghead{Zhou et al.}{DISCO for Obj. Det. with Noisy Bounding Boxes}

\begin{document}

\maketitle

\begin{abstract}
Large-scale well-annotated datasets are of great importance for training an effective object detector.
However, obtaining accurate bounding box annotations is laborious and demanding.
Unfortunately, the resultant noisy bounding boxes could cause corrupt supervision signals and thus diminish detection performance.   
Motivated by the observation that the real ground-truth is usually situated in the aggregation region of the proposals assigned to a noisy ground-truth,
we propose DIStribution-aware CalibratiOn (DISCO) to model the spatial distribution of proposals for calibrating supervision signals.
In DISCO, spatial distribution modeling is performed to statistically extract the potential locations of objects.
Based on the modeled distribution, three distribution-aware techniques, \textit{i.e.}, distribution-aware proposal augmentation (DA-Aug), distribution-aware box refinement (DA-Ref), and distribution-aware confidence estimation (DA-Est), are developed to improve classification, localization, and interpretability, respectively.
Extensive experiments demonstrate that DISCO can achieve SOTA performance in this task, especially at high noise levels.
Code is available at \href{https://github.com/Correr-Zhou/DISCO}{https://github.com/Correr-Zhou/DISCO}.
\end{abstract}

\section{Introduction}

Object detection has made substantial progress in recent years \cite{ren2015faster,lin2017focal,carion2020end,sun2021sparse,zou2023object}, which is largely attributed to the utilization of large-scale well-annotated datasets \cite{everingham2010pascal,lin2014microsoft}.
However, obtaining accurate bounding box annotations is labor-intensive and demanding, especially for some real-world scenarios \cite{luo2021oxnet,chai2023deep, michaelis2019benchmarking,mao20223d}.
Specifically, inherent ambiguities of bounding boxes are often caused by object occlusion or unclear boundaries \cite{he2019bounding}.
Moreover, insufficient domain expertise and the strenuous workload can also lead to low-quality labeling of bounding boxes \cite{liu2022robust}. 
In practical deployments and applications, vanilla object detectors will inevitably suffer from such noisy bounding box annotations.
Therefore, it is of scientific interest to explore how to tackle noisy bounding boxes in object detection.

\begin{figure}[t]
    \centering
    \includegraphics[width=1\linewidth]{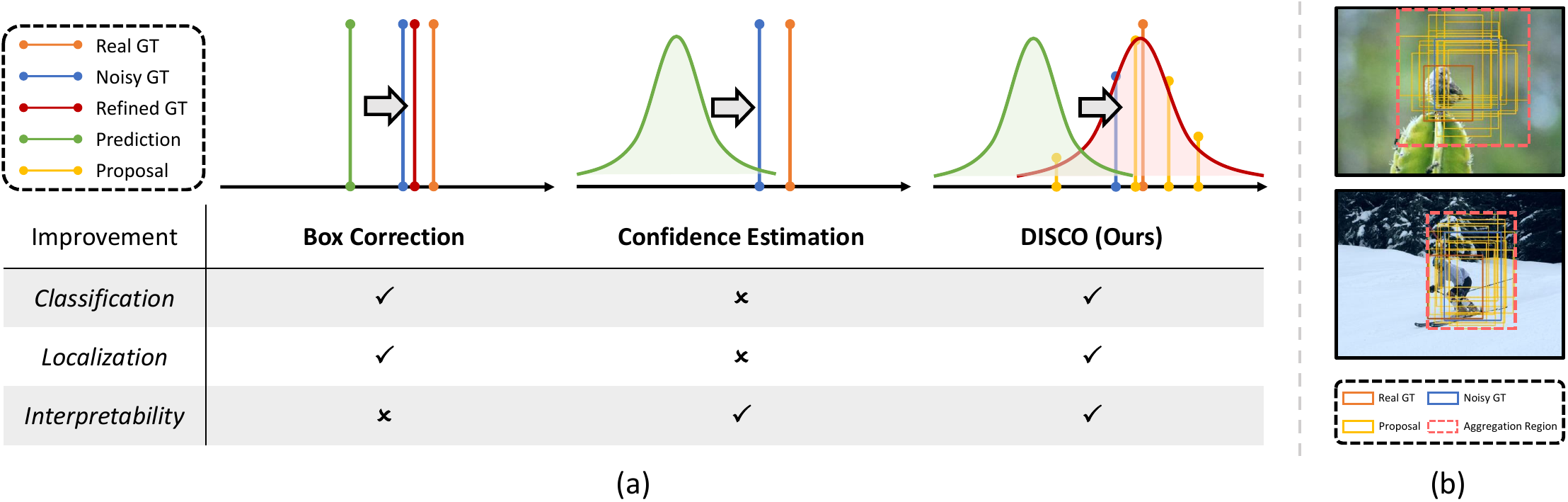}
    \vspace{-6mm}
    \caption{(a) \textbf{Trait comparison of existing solutions and our DISCO.}
        Their learning behaviors for one single border of bounding boxes are presented above. 
        (b) \textbf{Proposal aggregation in object detection with noisy bounding boxes.}
        The real ground-truth is usually situated in the aggregation region of the proposals assigned to a noisy ground-truth. 
    }
    \vspace{-3mm}
    \label{fig:task}
\end{figure}

Due to the degenerated supervision introduced by noisy annotations, object detection with noisy bounding boxes remains a challenging problem.
Obviously, such corrupt supervision signals could weaken the localization precision of object detectors.
Besides, although the classification accuracy is less affected \cite{liu2022robust}, noisy bounding boxes do introduce biased category features during training, which reduces the generalization capability of classification.
Notably, there is also a significant concern about the lack of interpretability for box predictions, especially considering the influence of noisy bounding box annotations.

Encountering the above challenges, existing solutions still exhibit drawbacks (see Figure~\ref{fig:task}(a)).
Several methods are dedicated to correcting noisy bounding boxes \cite{liu2022robust,li2020towards,xu2021training}, aiming to mitigate the effect of noise.
However, their performance gains are constrained by heuristic box correction approaches, 
and the detector cannot identify which bounding boxes are inaccurately predicted.
An alternative series of methods focuses on equipping the model with the ability to estimate the confidence of predicted bounding boxes \cite{he2019bounding,li2020generalized,choi2019gaussian}, by which the detector can be more robust against noisy box annotations. 
Despite these efforts, it is unfortunate that the detector is still plagued by flawed supervision during training.

Essentially, the corrupt supervision signals should be blamed for the above issues. 
In this work, we expect to answer the following question: \textit{How to properly calibrate the corrupt supervision signals?}
As shown in Figure~\ref{fig:task}(b), we observed that the real ground-truth is usually situated in the aggregation region of the proposals assigned to a noisy ground-truth, 
showing that the spatial distribution of proposals can act as a statistical prior for the potential locations of objects.
Thus, we propose \textit{DIStribution-aware CalibratiOn} (\textit{DISCO}) to model the spatial \textit{distribution} of proposals for \textit{calibrating} supervision signals. 
For each group of the assigned proposals, we perform \textit{spatial distribution modeling}, 
statistically extracting potential locations of objects.
Based on the modeled distribution, we develop three distribution-aware techniques to improve \textit{classification}, \textit{localization}, and \textit{interpretability}, respectively:
1) \textit{Distribution-aware proposal augmentation} (\textit{DA-Aug}):
Additional proposals are generated from the distribution to enrich category features, and then the proposal with the highest classification score is collected to boost classification performance;
2) \textit{Distribution-aware box refinement} (\textit{DA-Ref}): 
With a non-linear weighting strategy, noisy ground-truth is fused with the distribution into a refined ground-truth to achieve superior bounding box regression;
3) \textit{Distribution-aware confidence estimation} (\textit{DA-Est}): 
An extra estimator is integrated into the detection head, with the distribution variance elegantly acting as its supervision, to estimate the confidence of box predictions.
Our experiments show that DISCO can attain SOTA performance on large-scale noisy detection datasets, especially at high noise levels.

Our main contributions are listed as follows:
(1) Motivated by the observation about proposal aggregation, we propose DISCO to calibrate supervision signals with spatial distribution modeling.
(2) To improve classification, localization, and interpretability, we introduce three techniques (\textit{i.e.}, DA-Aug, DA-Ref, and DA-Est) to collaborate with the modeled distribution in a distribution-aware manner.
(3) Comprehensive experiments show that DISCO can attain SOTA performance and achieve satisfactory interpretability for box predictions.


\section{Related Works}

\textbf{Object Detection.}
The goal of object detection is to recognize what objects are present and where they are situated.
Faster-RCNN \cite{ren2015faster} is a classic detection framework with a two-stage strategy, and is widely adopted and improved in subsequent works \cite{cai2018cascade,pang2019libra,sun2021sparse}. 
Moreover, RetinaNet \cite{lin2017focal}, YOLO \cite{redmon2016you}, and CenterNet \cite{zhou2019objects} delve into strengthening the performance of one-stage detectors.
Recently, transformer-based detectors \cite{carion2020end,zhu2020deformable,zhang2022dino} also attracted the attention of the community, which conducts object detection in an end-to-end fashion.
Trained with accurate box annotations, object detectors can achieve satisfactory performance.
However, object detection with noisy bounding boxes remains an under-explored subproblem.

\noindent \textbf{Object Detection with Noisy Annotations.}
Noisy annotations of an object detection dataset comprise noisy category labels and noisy bounding boxes.
Previous works \cite{chadwick2019training,li2020towards,xu2021training} have jointly handled these two types of noisy annotations.
Unlike this setting, we focus on training an object detector with noisy bounding boxes, since box noise is more common and challenging in realistic scenarios \cite{liu2022robust}.
The SOTA method for this tough task is called OA-MIL \cite{liu2022robust}, which adopts a multi-instance learning (MIL) framework at the object level to correct bounding boxes.
Besides, some approaches aiming to boost the robustness of detectors, such as KL Loss \cite{he2019bounding}, can also contribute to performance improvement in this task. 

\noindent \textbf{Weakly Supervised Object Detection.}
Weakly supervised object detection (WSOD) is also a relevant task, where only image-level labels can be accessed to train an object detector.
The mainstream solution is to treat WSOD as a MIL problem \cite{bilen2016weakly,wan2018min,chen2020slv,tang2017multiple}, where each training image is constructed as a bag of instances.
To handle the non-convex optimization of MIL, spatial regularization \cite{diba2017weakly,wan2018min}, optimization strategy \cite{tang2017multiple,wan2019c}, and context information \cite{kantorov2016contextlocnet,wei2018ts2c} are introduced to attain better convergence.
Moreover, it is worth noting that SD-LocNet \cite{zhang2019learning} contributes a self-directed optimization strategy to handle object instances with noisy initial locations.
Unfortunately, WSOD always results in relatively inaccurate box predictions due to the lack of fine-grained supervision.
Effective methods for object detection with noisy bounding boxes can contribute to further refining these box predictions.


\begin{figure*}[t]
    \centering
    \includegraphics[width=1\linewidth]{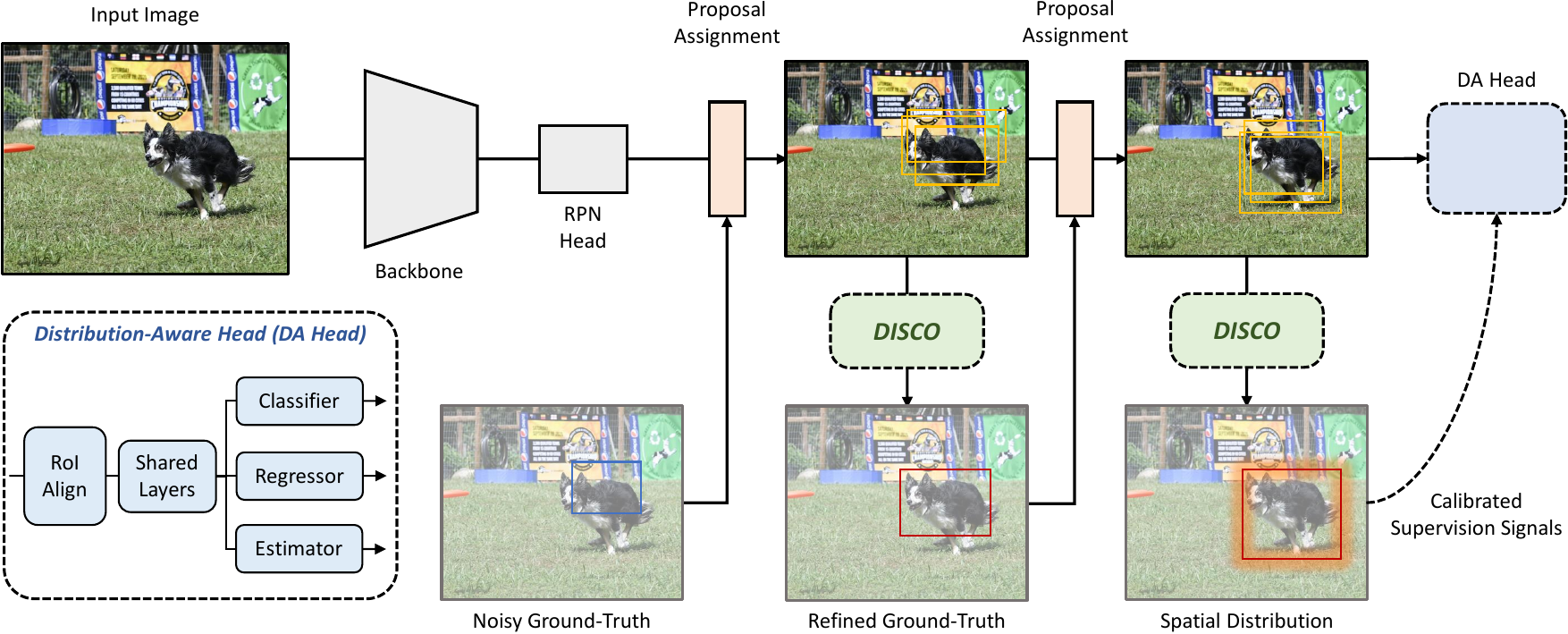}
    \vspace{-6mm}
    \caption{\textbf{Training pipeline with DISCO.}
        In DISCO, spatial distribution modeling (Section~\ref{sec:SDM}) is performed firstly, followed by three distribution-aware techniques, \textit{i.e.}, DA-Aug (Section~\ref{sec:DA-Aug}), DA-Ref (Section~\ref{sec:DA-Ref}), and DA-Est (Section~\ref{sec:DA-Est}), to collaborate with the modeled distribution.
        Note that we additionally integrate an estimator into the vanilla detection head to construct the distribution-aware head (DA head) for the implementation of DISCO.
    }
    \vspace{-2mm}
    \label{fig:pipline}
\end{figure*}

\section{Methodology}
DISCO is a training-time approach designed for two-stage detectors.
In a training iteration, it is performed \textit{twice} using the distribution-aware head (DA head) with the assigned proposals as input (see Figure~\ref{fig:pipline}).
Two times of DISCO follow the same process and have only subtle differences for different purposes. 
The first time aims to yield a refined ground-truth for proposal re-assignment, by which better-matching proposals can be obtained.
The second time aims to produce spatial distributions of proposals acting as superior supervision.
In the following, we start by describing spatial distribution modeling in Section~\ref{sec:SDM}.
Then, we will introduce DA-Aug, DA-Ref, and DA-Est in Section~\ref{sec:DA-Aug}, \ref{sec:DA-Ref} and \ref{sec:DA-Est}, respectively, in which we will also detail the differences between these two times of DISCO.

\subsection{Spatial Distribution Modeling}
\label{sec:SDM}
\begin{figure*}[t]
    \centering
    \includegraphics[width=0.95\linewidth]{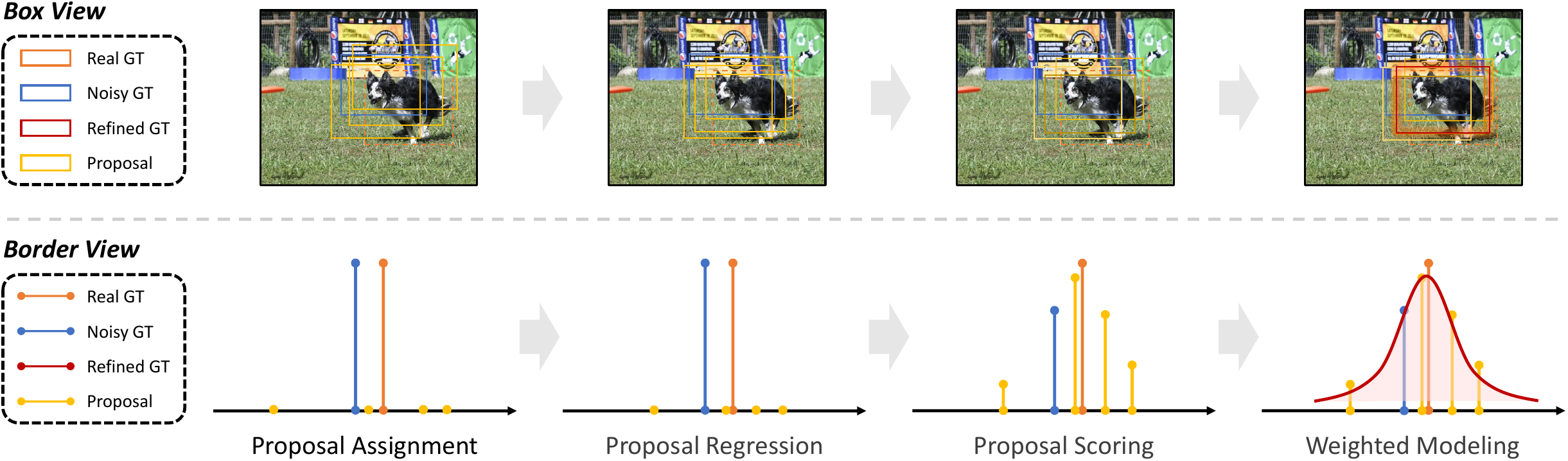}
    \vspace{-2mm}
    \caption{\textbf{Illustration of spatial distribution modeling.}
        For clarity, we present the process in the view of the whole box and one single border.
        Note that the length of the vertical line indicates its weight.
        Here the refined ground-truth is essentially a spatial distribution.
    }
    \vspace{-4mm}
    \label{fig:SPM}
\end{figure*}

In DISCO, spatial distribution modeling is conducted for each group of the proposals assigned to a noisy/refined ground-truth (see Figure~\ref{fig:SPM}).
Let $\mathbf{P}^i = [P^i_1, P^i_2, ..., P^i_{N^i}] \in \mathbb{R}^{N^i \times 4}$ denotes the $i$-th group of the proposals, where  $N^i$ is the number of the proposals in $\mathbf{P}^i$.
Moreover, $\mathbf{P}^i$ is associated with a category indicator $l^i \in \{1, 2, ..., L\}$ where $L$ is the number of categories.
Note that the noisy ground-truth is included in $\mathbf{P}^i$ as commonly done,
and each proposal $P^i_j \in \mathbb{R}^4$ represents four coordinates of bounding boxes. 
First, the features of the proposals  in $\mathbf{P}^i$ are extracted as
\begin{align}
     \mathbf{F}^{i} = \mathcal{F}(\mathbf{P}^{i}, \mathbf{X}) = [F^i_1, F^i_2, ..., F^i_{N^i}] \in \mathbb{R}^{N^i \times D},
\end{align}
where $\mathcal{F}(\cdot, \cdot)$ is the joint operation of RoIAlign \cite{he2017mask} and two shared fully-connected layers, and $\mathbf{X}$ is the feature maps produced by the backbone.
Each proposal $P^i_j$ corresponds to a $D$-dimensional feature vector $F^i_j$.
Then, we adopt the regressor $\mathcal{R}(\cdot)$ of the DA head to predict proposal offsets for further localization and update the features, which is formulated as
\begin{equation}
     \mathbf{P}^{i*} = \operatorname{Trans}(\mathbf{P}^i, \mathcal{R}(\mathbf{F}^i)) \in \mathbb{R}^{N^i \times 4},
     \quad
      \mathbf{F}^{i*} = \mathcal{F}(\mathbf{P}^{i*}, \mathbf{X}) \in \mathbb{R}^{N^i \times D},
\end{equation}
where $\operatorname{Trans(\cdot, \cdot)}$ is a function that translates predicted offsets to proposals \cite{ren2015faster}. 
Following \cite{liu2022robust}, we utilize classification scores to measure the possibilities of object locations.
Therefore, the classifier $\mathcal{C}(\cdot)$ is used to score the proposals $\mathbf{F}^{i*}$, which is defined as
\begin{equation}
    \mathbf{S}^{i} = \mathcal{C}(\mathbf{F}^{i*}) \in \mathbb{R}^{N^i \times (L+1)},
    \quad
    S^{i} = \operatorname{LookUp}(\mathbf{S}^{i}, l^i) = [s^{i}_1, s^{i}_2, ..., s^{i}_{N^i}] \in \mathbb{R}^{N^i},
    \label{eq:cls_score} 
\end{equation}
where $\operatorname{LookUp}(\cdot, \cdot)$ is a look-up operation that extracts the $l^i$-th column from $\mathbf{S}^{i}$, and the resultant $S^{i}$ denotes the classification scores of the corresponding category.
Subsequently, we utilize $S^{i}$ to produce the normalized weights $W^{i}$ for the proposals $\mathbf{F}^{i*}$, expressed as
\begin{equation}
    W^{i} = \operatorname{Softmax}(S^{i}, T)  = [w^{i}_1, w^{i}_2, ..., w^{i}_{N^i}] \in \mathbb{R}^{N^i},
    \label{eq:weight}
\end{equation}
where $\operatorname{Softmax}(\cdot, \cdot)$ is the Softmax function to obtain normalized weights (sum up to $1$) and $T$ is the temperature coefficient \cite{hinton2015distilling}. 
Finally, we model the spatial distribution of $\mathbf{P}^{i*}$ as a four-dimensional Gaussian distribution by directly calculating its parameters (\textit{i.e.}, mean $\mu^{i}$ and standard deviation $\sigma^{i}$) in a weighting manner, which can be formulated as
\begin{equation}
    \mu^{i} = \sum_{j=1}^{N^i} w^{i}_j * P^{i*}_j \in \mathbb{R}^{4},
    \quad
    \sigma^{i} = \sqrt{\sum_{j=1}^{N^i} w^{i}_j * (P^{i}_j - \mu^{i})^2} \in \mathbb{R}^{4},
    \label{eq:mu}
\end{equation}
where we assume that each dimension of this Gaussian distribution is uncorrelated so that its standard deviation can be formulated as a four-dimensional vector.
This assumption has also been widely adopted in previous works \cite{he2019bounding, li2020generalized, choi2019gaussian}.
as it can simplify the modeling problem, make the method more computationally efficient, and is not detrimental to performance.

\subsection{Distribution-Aware Proposal Augmentation}
\label{sec:DA-Aug}
Instead of using heuristic approaches such as selective search \cite{uijlings2013selective} and edge box \cite{zitnick2014edge}, we propose to augment proposals with the modeled distribution, aiming to statistically cover more potential locations of objects.
Firstly, we create a Guassian noise matrix $\mathbf{G} \in \mathbb{R}^{N^\prime \times 4}$ whose each element is sampled from $\mathcal{N}(0, 1)$ to ensure randomness. 
Here $N^\prime$ is a hyperparameter that indicates the number of augmented proposals.
Then, augmented proposals $\mathbf{P}^{i\prime}$ can be generated by
\begin{equation}
    \mathbf{P}^{i\prime} = \mu^{i} + \mathbf{G} \odot \sigma^{i} = [P^{i\prime}_1, P^{i\prime}_2, ..., P^{i\prime}_{N^\prime}] \in \mathbb{R}^{N^\prime \times 4},
    \label{eq:aug}
\end{equation}
where $\odot$ denotes element-wise multiplication and the operations here are all conducted in a broadcasting fashion.
Following Equation~\ref{eq:cls_score}, its classification scores $S^{i\prime}$ can be obtained.
Subsequently, we incorporate the augmented proposals $\mathbf{P}^{i\prime}$ into $\mathbf{P}^{i}$, expressed as
\begin{equation}
    \mathbf{P}^{i*} \leftarrow \mathbf{P}^{i*} \oplus \mathbf{P}^{i\prime} \in \mathbb{R}^{(N^i + N^\prime) \times 4},
    \quad
    S^{i} \leftarrow S^{i} \oplus S^{i\prime} \in \mathbb{R}^{N^i + N^\prime},
\end{equation}
where $\oplus$ indicates proposal-wise concatenation.
To boost classification performance, the proposals with the highest classification score $s^{i}_\star = \max_j s^{i}_{j}$, which could contain representative category features, are collected to form a loss term for classification, formulated as
\begin{equation}
    \mathcal{L}_{\text{Aug}} = - \frac{1}{M} \sum_{i=1}^M \log(s^{i}_\star),
\end{equation}
where $M$ is the number of proposal groups in a batch. 
\textit{Note that} $\mathcal{L}_{\text{Aug}}$ \textit{does not be computed in the first-time DISCO}.
Finally, after proposal augmentation, we follow Equation~\ref{eq:mu} to model the spatial distribution of $\mathbf{P}^{i*}$ once again for a better representation of these proposals.

\subsection{Distribution-Aware Box Refinement}
\label{sec:DA-Ref}
As we mentioned before, the modeled spatial distribution can be considered as a statistical prior for the potential locations of objects. 
Therefore, it can act as guidance for noisy bounding box refinement.
First, we treat $\mu^{i}$ as a proposal, extract its feature with $\mathcal{F}(\cdot, \cdot)$, and adopt the classifier $\mathcal{C}(\cdot)$ to obtain its classification score $s^{i}_\mu$.
Then, the noisy bounding box $B^i$ is refined as $B^{i\prime}$ by a fusion strategy as
\begin{equation}
    B^{i\prime} = \phi(s^{i}_\mu) \cdot \mu^{i} + (1 - \phi(s^{i}_\mu)) \cdot B^i,
    \label{eq:fusion}
\end{equation}
where $\phi(\cdot)$ is a non-linear weighting function.
To stabilize the early stage of training, the fusion strategy is conditional on the classification score $s^{i}_\mu$, and thus the $\phi(\cdot)$ is defined as
\begin{equation}
    \phi(s^{i}_\mu) = \min((s^{i}_\mu)^\alpha, \beta),
    \label{eq:nonlinear}
\end{equation}
where $\alpha$ and $\beta$ are two hyperparameters.
It means that the higher $s^{i}_\mu$ is, the more the model would rely on $\mu^{i}$.
Finally, the refined $B^{i\prime}$ is used as supervision for the original proposals $\mathbf{P}^i$ to compute regression loss as
\begin{equation}
    \mathcal{L}_{\text{reg}} = \frac{1}{M} \sum_{i=1}^M \frac{1}{N^i} \sum_{j=1}^{N^i} \operatorname{Dist}(P^i_j, B^{i\prime}),
\end{equation}
where $\operatorname{Dist}(\cdot, \cdot)$ is a predefined distance function for two bounding boxes \cite{ren2015faster}.
\textit{It is worth noting that the first-time DISCO is ended with Equation~\ref{eq:fusion}} to obtain the refined ground-truth $B^{i\prime}$ for proposal re-assignment.

\subsection{Distribution-Aware Confidence Estimation}
\label{sec:DA-Est}
To estimate the confidence of predicted bounding boxes, we integrate an estimator $\mathcal{E}(\cdot)$ into the DA head. Note that $\mathcal{E}(\cdot)$ comprises only one fully connected layer. As excepted, the confidence $\mathbf{V}^i$ for the proposals $\mathbf{P}^i$ can be produced as
\begin{equation}
    \mathbf{V}^i = \mathcal{E}(\mathbf{F}^i) = [V^i_1, V^i_2, ..., V^i_{N^i}] \in \mathbb{R}^{N^i \times 4}.
\end{equation}
In the modeled spatial distribution, the variance (or standard deviation) can measure the border-wise variability of the potential locations of objects.
Therefore, the distribution variance can be elegantly adopted as the supervision of the estimator $\mathcal{E}(\cdot)$. 
The loss for training $\mathcal{E}(\cdot)$ is formulated as
\begin{equation}
    \mathcal{L}_{\text{Est}} = \frac{1}{M} \sum_{i=1}^M \frac{1}{N^i} \sum_{j=1}^{N^i} \|V^i_j - (\sigma^i)^2\|_1.
\end{equation}
Different from \cite{he2019bounding}, we train the estimator with direct supervision of variance rather than implicit supervision of bounding boxes.
Then, the estimated confidence (\textit{i.e.}, predicted variance) is used in Softer-NMS \cite{he2019bounding} for a better inference-time process. 
Finally, the overall loss function is formed as
\begin{equation}
    \mathcal{L}_{\text{All}} = \mathcal{L}_{\text{Cls}} + \mathcal{L}_{\text{Reg}} + \gamma \mathcal{L}_{\text{Est}} + \lambda \mathcal{L}_{\text{Aug}},
    \label{eq:overall}
\end{equation}
where $\gamma$ and $\lambda$ is two hyperparameters to down-weight $\mathcal{L}_{\text{Est}}$ and $\mathcal{L}_{\text{Aug}}$ respectively, and $\mathcal{L}_{\text{Cls}}$ is a cross-entropy classification loss for the original proposals $\mathbf{P}^i$ \cite{ren2015faster}.


\begin{table*}[t]
  \centering
  \setlength{\tabcolsep}{1pt}
  \renewcommand{\arraystretch}{1.6}
  \resizebox{\linewidth}{!}{
    \setlength{\tabcolsep}{2mm}{
\begin{tabular}{l|cccc|cccccc|cccccc}
\specialrule{0.15em}{0pt}{0pt}
\multirow{3}[4]{*}{Method} & \multicolumn{4}{c|}{VOC}      & \multicolumn{12}{c}{COCO} \bigstrut\\
\cline{2-17}      & \multicolumn{4}{c|}{Noise Level} & \multicolumn{6}{c|}{20\% Noise Level}     & \multicolumn{6}{c}{40\% Noise Level} \bigstrut[t]\\
      & 10\%  & 20\%  & 30\%  & 40\%  & $\text{AP}$ & $\text{AP}_{50}$ & $\text{AP}_{75}$ & $\text{AP}_\text{S}$ & $\text{AP}_\text{M}$ & $\text{AP}_\text{L}$ & $\text{AP}$ & $\text{AP}_{50}$ & $\text{AP}_{75}$ & $\text{AP}_\text{S}$ & $\text{AP}_\text{M}$ & $\text{AP}_\text{L}$ \bigstrut[b]\\
\hline
\hline
Clean-FasterRCNN \cite{ren2015faster} & 77.2  & 77.2  & 77.2  & 77.2  & 37.9  & 58.1  & 40.9  & 21.6  & 41.6  & 48.7  & 37.9  & 58.1  & 40.9  & 21.6  & 41.6  & 48.7 \bigstrut[t]\\
\hline
FasterRCNN \cite{ren2015faster} & 76.3  & 71.2  & 60.1  & 42.5  & 30.4  & 54.3  & 31.4  & 17.4  & 33.9  & 38.7  & 10.3  & 28.9  & 3.3   & 5.7   & 11.8  & 15.1 \bigstrut[t]\\
RetinaNet \cite{lin2017focal} & 71.5  & 67.5  & 57.9  & 45.0  & 30.0  & 53.1  & 30.8  & 17.9  & 33.7  & 38.2  & 13.3  & 33.6  & 5.7   & 8.4   & 15.9  & 18.0 \\
Co-teaching \cite{han2018co} & 75.4  & 70.6  & 60.9  & 43.7  & 30.5  & 54.9  & 30.5  & 17.3  & 34.0  & 39.1  & 11.5  & 31.4  & 4.2   & 6.4   & 13.1  & 16.4 \\
SD-LocNet \cite{zhang2019learning} & 75.7  & 71.5  & 60.8  & 43.9  & 30.0  & 54.5  & 30.3  & 17.5  & 33.6  & 38.7  & 11.3  & 30.3  & 4.3   & 6.0   & 12.7  & 16.6 \\
FreeAnchor \cite{zhang2019freeanchor} & 73.0  & 67.5  & 56.2  & 41.6  & 28.6  & 53.1  & 28.5  & 16.6  & 32.2  & 37.0  & 10.4  & 28.9  & 3.3   & 5.8   & 12.1  & 14.9 \\
KL Loss \cite{he2019bounding} & 75.8  & 72.7  & 64.6  & 48.6  & 31.0  & 54.3  & 32.4  & 18.0  & 34.9  & 39.5  & 12.1  & 36.7  & 3.7   & 6.2   & 13.0  & 17.4 \\
OA-MIL \cite{liu2022robust} & 77.4 & 74.3  & 70.6  & 63.8  & 32.1  & \textbf{55.3} & 33.2  & 18.1  & 35.8  & \textbf{41.6} & 18.6  & 42.6  & 12.9  & 9.2   & 19.0  & 26.5 \\
DISCO (Ours) & \textbf{77.5}  & \textbf{75.3} & \textbf{72.1} & \textbf{68.7} & \textbf{32.3} & 54.7  & \textbf{34.5} & \textbf{18.7} & \textbf{35.8} & 41.2  & \textbf{21.2} & \textbf{45.7} & \textbf{16.9} & \textbf{11.4} & \textbf{24.7} & \textbf{27.8} \bigstrut[b]\\
\specialrule{0.15em}{0pt}{0pt}
\end{tabular}%
}}   
     \caption{\textbf{Benchmark results on VOC and COCO.}
     Note that Clean-FasterRCNN is trained with clean annotations for reference.
     In noisy settings, the best results are marked in bold. 
     }
    \vspace{-2mm}
    \label{tab:benchmark}
\end{table*}

\section{Experiments}
\label{sec:exp}

To ensure fairness, our experimental setup is all aligned with \cite{liu2022robust}.
Specifically, two large-scale detection datasets, \textit{i.e.}, Pascal VOC 2007 (VOC) \cite{everingham2010pascal} and MS-COCO 2017 (COCO) \cite{lin2014microsoft}, are adopted in the experiments.
Noisy box annotations are simulated by perturbing clean ones at various noise levels which are set to $\{10\%, 20\%, 30\%, 40\%\}$ for VOC and $\{20\%, 40\%\}$ for COCO.
We report $\text{AP}_{50}$ for VOC and $\{\text{AP}, \text{AP}_{50}, \text{AP}_{75}, \text{AP}_{S}, \text{AP}_{M}, \text{AP}_{L}\}$ for COCO.
Due to the space limit, more details are included in Appendix~\ref{app:setup}.

\begin{table}[t]
  \centering
  \setlength{\tabcolsep}{1pt}
  \renewcommand{\arraystretch}{1.6}
    \begin{minipage}[c]{0.48\linewidth}
    \centering
        \resizebox{\linewidth}{!}{
        \setlength{\tabcolsep}{1.5mm}{
        \begin{tabular}{l|l|cccccc}
        \hline
        Method & Noise Level & $\text{AP}$ & $\text{AP}_{50}$ & $\text{AP}_{75}$ & $\text{AP}_\text{S}$ & $\text{AP}_\text{M}$ & $\text{AP}_\text{L}$ \bigstrut\\
        \hline
        \hline
        OA-MIL \cite{liu2022robust} & \multirow{2}[2]{*}{10\%} & 35.1  & 57.2  & 37.9  & 20.5  & 38.5  & 44.9 \bigstrut[t]\\
        DISCO (Ours) &       & \textbf{36.1} & \textbf{57.3} & \textbf{39.4} & \textbf{20.8} & \textbf{39.5} & \textbf{45.7} \bigstrut[b]\\
        \hline
        OA-MIL \cite{liu2022robust} & \multirow{2}[2]{*}{30\%} & 24.6  & 49.1  & 21.9  & 13.8  & 27.5  & 32.7 \bigstrut[t]\\
        DISCO (Ours) &       & \textbf{26.4} & \textbf{49.8} & \textbf{25.3} & \textbf{14.2} & \textbf{29.7} & \textbf{34.2} \bigstrut[b]\\
        \hline
        \end{tabular}%
        }}
    \end{minipage}
    \begin{minipage}[c]{0.48\linewidth}
    \centering
        \resizebox{\linewidth}{!}{
        \setlength{\tabcolsep}{2.4mm}{
        \begin{tabular}{l|c|cccccc}
        \hline
        \multirow{2}[4]{*}{Method} & VOC   & \multicolumn{6}{c}{COCO} \bigstrut\\
    \cline{2-8}          & $\text{AP}_{50}$ & $\text{AP}$ & $\text{AP}_{50}$ & $\text{AP}_{75}$ & $\text{AP}_\text{S}$ & $\text{AP}_\text{M}$ & $\text{AP}_\text{L}$ \bigstrut\\
        \hline
        \hline
        OA-MIL \cite{liu2022robust} & 77.1  & 37.0  & 57.9  & 40.3  & 21.8  & 40.6  & 47.6 \bigstrut[t]\\
        DISCO (Ours) & \textbf{78.0} & \textbf{38.0} & \textbf{57.9}  & \textbf{41.9} & \textbf{21.9} & \textbf{41.4}  & \textbf{48.5} \bigstrut[b]\\
        \hline
        \end{tabular}%
        }}
    \end{minipage}
     \caption{Left:
     \textbf{Experimental results on COCO at 10\% and 30\% noise levels.}
     DISCO can still achieve SOTA performances.
     Right:
     \textbf{Experimental results on the original VOC and COCO.}
     DISCO can provide performance improvement without manually introducing noise.
     }
    \vspace{-2mm}
    \label{tab:coco_add}
\end{table}

\subsection{Results and Discussions}
We compare DISCO with the SOTA methods of this task, including FasterRCNN \cite{ren2015faster}, Co-teaching \cite{han2018co}, SD-LocNet \cite{zhang2019learning}, KL Loss \cite{he2019bounding}, and OA-MIL \cite{liu2022robust}. Besides, the results of two one-stage methods are presented for a further comparison, including RetinaNet \cite{lin2017focal} and FreeAnchor \cite{zhang2019freeanchor}.
For reference, we also report the result of Clean-FasterRCNN, which is trained with clean annotations under the same setup.

\noindent \textbf{Benchmark Results.}
Benchmark results are reported in Table~\ref{tab:benchmark}.
It can be observed that noisy bounding box annotations significantly reduce the performance of vanilla object detectors like FasterRCNN, especially at high box noise levels.
Moreover, Co-teaching and SD-LocNet can only marginally improve detection performance, showing that small-loss sample selection and sample weight assignment are not decent solutions for handling noisy bounding boxes.
Besides, even with better label assignment, FreeAnchor still underperforms in such a challenging task.
It is worth noting that KL Loss is a competitive method that also improves the interpretability of detectors.
Moreover, OA-MIL adopts a MIL-based training strategy by iteratively constructing object-level bags, attaining better detection performance than the aforementioned methods.
Our DISCO, which aims to calibrate the corrupt supervision signals caused by noisy bounding boxes, achieves SOTA performance on these two benchmarks.
Notably, it can significantly outperform the existing methods at high noise levels (i.e., $30\%$ and $40\%$), showing that our method is more robust to noisy bounding boxes.
Specifically, compared with OA-MIL, DISCO attains $+1.5\text{AP}_{50}$ and $+4.9\text{AP}_{50}$ improvement on VOC at the $30\%$ and $40\%$ noise levels respectively.
DISCO can also achieve $+2.6\text{AP}$, $+3.1\text{AP}_{50}$, and $+4.0\text{AP}_{75}$ improvement on COCO at the $40\%$ noise level. 

\begin{table*}[t]
  \centering
  \setlength{\tabcolsep}{1pt}
  \renewcommand{\arraystretch}{1.6}
  \resizebox{\linewidth}{!}{
    \setlength{\tabcolsep}{1.5mm}{

\begin{tabular}{ccc|cccccccccccccccccccc|c}
\specialrule{0.15em}{0pt}{0pt}
\cline{1-23}\multicolumn{3}{c|}{Component} & \multicolumn{20}{c|}{Category}                                                                                                                                & \multirow{2}[2]{*}{All} \bigstrut[t]\\
DA-Aug & DA-Ref & DA-Est & Aero  & Bicy  & Bird  & Boat  & Bot   & Bus   & Car   & Cat   & Cha   & Cow   & Dtab  & Dog   & Hors  & Mbik  & Pers  & Plnt  & She   & Sofa  & Trai  & Tv    &  \bigstrut[b]\\
\hline
\hline
$\checkmark$ &       &       & 49.4  & 69.5  & 47.4  & 32.1  & 35.2  & 62.3  & 64.1  & 60.3  & 31.9  & 55.1  & 41.5  & 61.8  & 54.3  & 56.8  & 58.7  & 22.5  & 48.6  & 49.7  & 49.8  & 51.3  & 50.1 \bigstrut[t]\\
      & $\checkmark$ &       & 56.0  & 70.6  & 56.0  & 38.5  & 33.0  & 64.7  & 74.8  & 77.4  & 32.2  & 58.5  & 42.4  & 72.1  & 65.6  & 64.8  & 62.5  & 23.4  & 51.2  & 51.5  & 65.7  & 50.5  & 55.6 \\
      & $\checkmark$ & $\checkmark$ & 61.2  & 74.4  & 59.7  & 43.1  & 37.0  & 69.4  & 75.2  & 73.3  & 34.8  & 64.1  & 54.5  & 74.1  & 71.7  & 66.0  & 66.7  & 28.7  & 54.1  & 55.4  & 70.5  & 60.2  & 59.7 \\
$\checkmark$ & $\checkmark$ &       & 69.9  & \textbf{77.1} & 68.2  & \textbf{47.2} & 49.9  &  70.9 & 80.6  & 80.8  & 43.0  & 76.4  & 60.0  & \textbf{82.6} & 81.0  & 74.4  & 73.4  & 39.2  & 62.7  & 64.3  & 67.9  & 68.6  & 66.9 \\
$\checkmark$ & $\checkmark$ & $\checkmark$ & \textbf{71.5} & 76.9  & \textbf{71.5} & 45.6  & \textbf{52.2} & \textbf{76.1} & \textbf{81.2} & \textbf{83.2} & \textbf{43.4} & \textbf{79.8} & \textbf{60.3} & 81.5  & \textbf{82.9} & \textbf{75.4} & \textbf{73.6} & \textbf{40.6} & \textbf{64.4} & \textbf{68.2} & \textbf{76.8} & \textbf{70.0} & \textbf{68.7} \bigstrut[b]\\
\specialrule{0.15em}{0pt}{0pt}
\end{tabular}

    }
}
     \caption{\textbf{Ablation studies of component effectiveness.} Note that per-category performance is reported for a detailed comparison. The proposed components (\textit{i.e.}, three distribute-aware techniques) of our DISCO can all contribute to performance improvement.
     }
    \vspace{-4mm}
    \label{tab:component}
\end{table*}

\noindent \textbf{Additional Evaluations.}
To further demonstrate the effectiveness of DISCO, We compare it with OA-MIL in more settings other than those included in \cite{liu2022robust}.
The additional evaluations include two aspects  (see Table~\ref{tab:coco_add}):
1) Performance on COCO at $10\%$ and $30\%$ noise levels:
Compared to OA-MIL, DISCO can still achieve SOTA performance in additional noisy settings on COCO, suggesting its flexibility for various noise levels;
2) Performance on the original VOC and COCO (\textit{i.e.}, the noise level is set to $0\%$): Without manually introducing noise, DISCO can still provide performance improvement on original datasets, especially on VOC, showing that it has the potential to be generalized to real-world noisy scenarios.
 
\begin{figure}[t]
    \centering
    \includegraphics[width=0.7\linewidth]{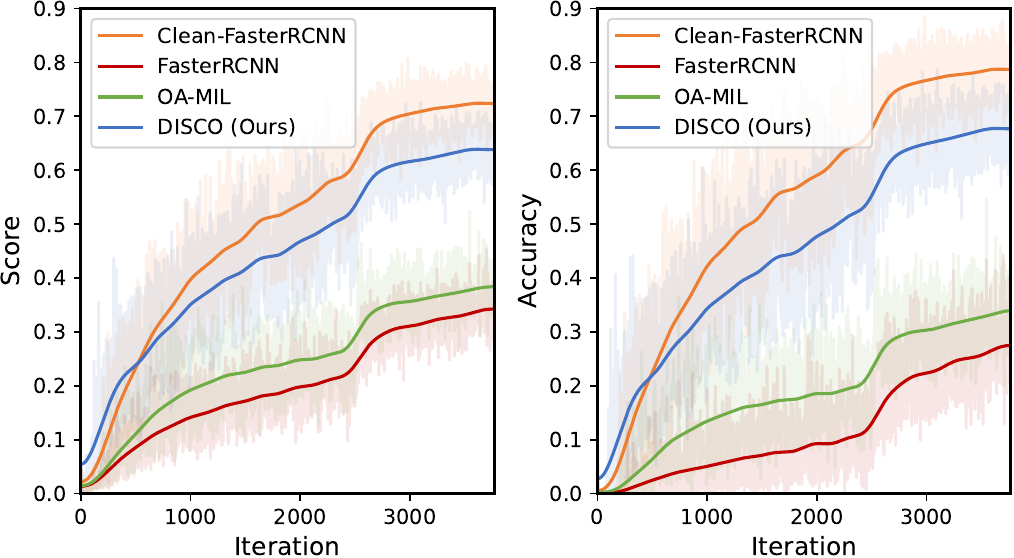}
    \caption{\textbf{Illustration of classification performance improvement.}
        Left: Average classification scores of the positive proposals for corresponding categories.
        Right: Classification accuracy of the positive proposals.
        Compared to OA-MIL, DISCO provides better improvement for classification, which even approaches the results of training with clean annotations.
    }
    \vspace{-4mm}
    \label{fig:classification}
\end{figure}

\subsection{Ablation Studies}

To investigate the effectiveness of three key techniques in DISCO, we gradually integrate them into training.
Note that the implementation of DA-Est is heavily based on DA-Ref thus it cannot be used independently.
The experimental results are reported in Table~\ref{tab:component}, which are based on VOC at the $40\%$ noise level.
Notably, using only DA-Aug or DA-Ref can considerably contribute to performance improvement. 
DA-Ref seems to be more effective since it comes with refined ground-truths for better localization.
Moreover, its detection performance can be further boosted when collaborating with DA-Aug or DA-Est.
It is also worth noting that DA-Est can achieve $+4.1 \text{AP}_{50}$ improvement by enhancing the robustness of detectors.
Using all three techniques, our DISCO can attain superior detection performance in almost all categories.
Because of the space limit, more ablation studies such as hyperparameter sensitivity and backbone compatibility are provided in Appendix~\ref{app:more_abl}.

\begin{figure}[t]
    \centering
    \includegraphics[width=1\linewidth]{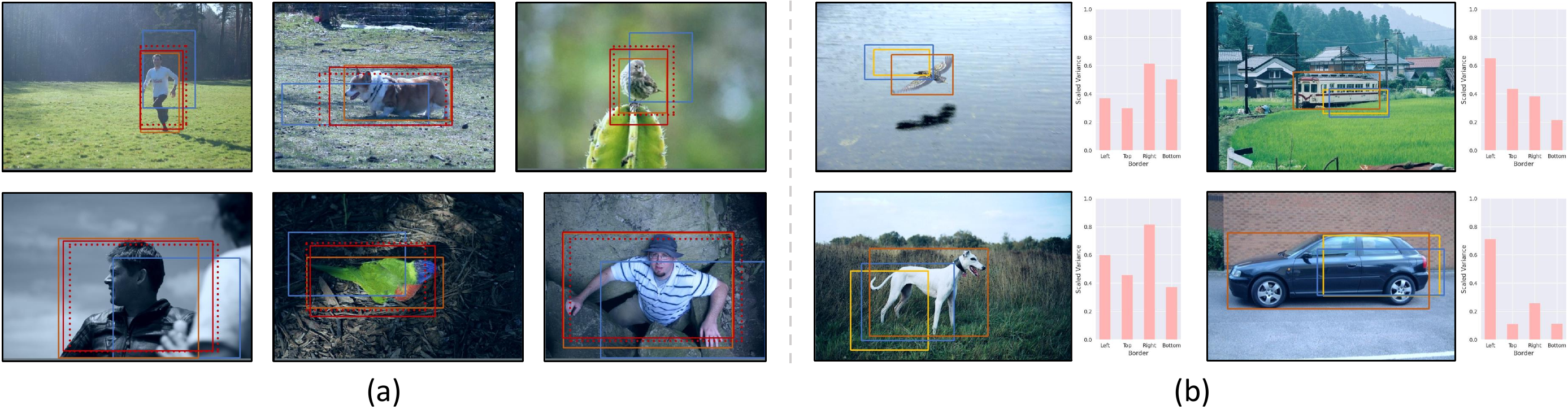}
    \vspace{-6mm}
    \caption{(a)
        \textbf{Qualitative results of box refinement in DISCO.}
        Real ground-truths and noisy ground-truths are marked in \textit{orange} and \textit{blue}.
        Refined bounding boxes produced by the first-/second-time DISCO are indicated in \textit{dotted/solid red}.
        The first-time refined boxes can cover the objects more tightly than noisy ground-truths, and the second-time refinement can further contribute to more precise ones.
        (b)
        \textbf{Qualitative results of interpretability in DISCO.}
        We randomly choose an assigned proposal (\textit{yellow}) per image to report its estimated variances.
        Real ground-truths and noisy ground-truths are marked in \textit{orange} and \textit{blue}.
        Note that the variance is scaled by the width and height for clarity.
        With the proposed DA-Est, DISCO can estimate reasonable variances for each border of box prediction.
    }
    \vspace{-4mm}
    \label{fig:localization}
\end{figure}

\subsection{Further Analysis}
In this subsection, additional evidence and discussion are provided to further analyze the advantages of DISCO in \textit{classification}, \textit{localization}, and \textit{interoperability}. 
Unless otherwise specified, the following experiments are all based on VOC at the $40\%$ noise level.

\noindent \textbf{Classification Performance Improvement.}
In DISCO, DA-Aug is used to generate proposals in the potential locations of objects for obtaining representative category features, by which classification performance can be boosted.
The evidence for advantage is provided in Figure~\ref{fig:classification}.
It shows that noisy bounding box annotations can badly reduce the classification scores and the accuracy of foreground features.
Notably, OA-MIL \cite{liu2022robust} can also enhance classification performance.
More importantly, DISCO provides superior improvement for classification, which even approaches the results of training with clean annotations.

\noindent \textbf{Box Refinement for Better Localization.}
To improve the localization capability of detectors, DA-Ref utilizes the modeled distributions of proposals for noisy box refinement (see Figure~\ref{fig:localization}(a)).
Note that DISCO is performed twice in a training iteration and thus there are two successive refined boxes, where the first one is for proposal re-assignment and the second one acts as the supervision for regression.
Since box refinement is very challenging when given only noisy ground-truths, it is natural that refined ones may be not exactly identical to real ones.
Even so, the refined boxes can cover the objects more tightly than noisy ground-truths.
Furthermore, the second-time refinement contributes to more precise ones,
showing the effectiveness of our refinement strategy.

\noindent \textbf{Interpretability with Confidence Estimation.}
We introduce interpretability into box predictions in DISCO, aiming to enhance the robustness of detectors.
The term "interpretability" is used to convey that our box predictions are interpretable, which is also adopted in \cite{he2019bounding}.
This is implemented by estimating the confidence of each border of the predicted bounding boxes with the variance of the modeled distribution as its supervision (see Figure~\ref{fig:localization}(b)).
For a predicted border that deviates largely from the real one, DISCO could estimate a relatively large variance, indicating low confidence for this prediction.  
Such a crucial property enhances the practicability of DISCO in realistic scenarios.


\section{Conclusions}
In this paper, we focus on an under-explored and challenging problem termed object detection with noisy bounding boxes.
Motivated by the observation about proposal aggregation, we propose DISCO to calibrate the corrupt supervision signals.
Spatial distribution modeling is performed and then three distribution-aware techniques (\textit{i.e.}, DA-Aug, DA-Ref, and DA-Est) are adopted successively.
Experiments show that DISCO can achieve SOTA performance.
We believe that DISCO can serve as a stronger baseline for this task and expect it can motivate more works in the fields of object detection and noisy label learning.

\section*{Acknowledgements}
The work described in this paper was supported in part by the grants from China National Key R\&D Program (Grant No. 2023YFE0202700, 2022YFE0200700), 
Hong Kong Innovation and Technology Fund (Project No. MHP/092/22),
NSFC General Project (62072452),
Regional Joint Fund of Guangdong (Grant No. 2021B1515120011),
National Natural Science Foundation of China (Project No. 6237073934, 3234101132), 
and Natural Science Foundation of Guangdong Province (2022A1515011579).
\bibliography{egbib}


\clearpage
\appendix

\section{Details of the Experimental Setup}
\label{app:setup}
\subsection{Datasets}
Two large-scale image datasets are adopted in our experiments, including Pascal VOC 2007 \cite{everingham2010pascal} and MS-COCO 2017 \cite{lin2014microsoft}.
Pascal VOC 2007 (VOC) is a standard dataset for object detection, consisting of $9,963$ images with $24,640$ box annotations.
MS-COCO 2017 (COCO) is also a popular object detection benchmark, containing $328,000$ images of generic objects.
Following \cite{liu2022robust}, clean annotations are perturbed to simulate noisy bounding box annotations in our experiments, which is performed once for each dataset.
Specifically, let $c_{\text{x}}, c_{\text{y}}, w, h$ represent the central x-axis coordinate, central y-axis coordinate, width, and height of a clean bounding box, respectively.
We simulate a noisy bounding box by randomly shifting and scaling a clean one, which can be formulated as
\begin{equation}
\left\{
\begin{aligned}
    & \hat{c}_{\text{x}} = c_{\text{x}} + \Delta_{\text{x}} \cdot w, \
    & \hat{c}_{\text{y}} = c_{\text{y}} + \Delta_{\text{y}} \cdot h, \\
    & \hat{w} = (1 + \Delta_{\text{w}}) \cdot w, \
    & \hat{h} = (1 + \Delta_{\text{h}}) \cdot h,
\end{aligned}
\right.
\label{eq:simulation}
\end{equation}
where $\Delta_{\text{x}}$, $\Delta_{\text{y}}$, $\Delta_{\text{w}}$, and $\Delta_{\text{h}}$ obey the uniform distribution $U(-n, n)$ and $n$ is the noise level.
For example, when $n$ is set to $40\%$, $\Delta_{\text{x}}$, $\Delta_{\text{y}}$, $\Delta_{\text{w}}$, and $\Delta_{\text{h}}$ would ranges from $-0.4$ to $0.4$.
Note that Equation~\ref{eq:simulation} is conducted on each bounding box of the training set.
Such a noise simulation can guarantee access to real ground-truths for analyzing training behaviors and evaluating the performance of box refinement.
Noise levels are set to $\{10\%, 20\%, 30\%, 40\%\}$ for VOC and $\{20\%, 40\%\}$ for COCO.

\subsection{Implementation Details}
Following \cite{liu2022robust}, we implement our method on FasterRCNN \cite{ren2015faster} with ResNet-50 \cite{he2016deep} as the backbone.
The idea of DISCO can be easily generalized to other frameworks and we choose to perform our experiments with FasterRCNN as it is widely adopted \cite{kaur2023comprehensive}.
As a common practice, the model is trained with the ``$1\times$" schedule \cite{girshick2018detectron}.
Notably, all other training configurations are aligned with \cite{liu2022robust} to ensure fairness.
As commonly done, mean average precision (\text{mAP}@$.5$) and mAP@$[.5, 95]$ are used for VOC and COCO respectively.
Specifically, we report $\text{AP}_{50}$ for VOC and $\{\text{AP}, \text{AP}_{50}, \text{AP}_{75}, \text{AP}_{S}, \text{AP}_{M}, \text{AP}_{L}\}$ for COCO.

\subsection{Hyperparameter Selections}
\label{app:hyper_select}
There are six hyperparameters in DISCO, including the temperature coefficient $T$, the augmented proposal number $N^\prime$, two box fusion hyperparameters $\alpha$ and $\beta$, and two loss weights $\gamma$ and $\lambda$. 
As there are no additional validation sets available, we tuned these hyperparameters based on the performance on the training set with clean annotations, which can also avoid the leakage of test data.
For the sake of simplicity, we empirically fix $N^\prime$ and $\gamma$ to $10$ and $0.3$, and then tuning $T \in [0.01, 0.2]$, $\alpha \in [3, 10]$, $\beta \in [0.7, 0.9]$, and $\lambda \in [0.01, 0.2]$.
To ensure reproducibility, the selected hyperparameters for all settings are reported in Table~\ref{tab:selection}.
Notably, we have just roughly tuned these hyperparameters by selecting some regular values, thus the performance of our method in Table~\ref{tab:benchmark} has the potential to be better.

\begin{table}[t]
  \centering
  \setlength{\tabcolsep}{1pt}
  \renewcommand{\arraystretch}{1.6}
  \resizebox{0.6\linewidth}{!}{
    \setlength{\tabcolsep}{3.4mm}{

\begin{tabular}{l|l|cccccc}
\specialrule{0.15em}{0pt}{0pt}
\multirow{2}[2]{*}{Dataset} & \multirow{2}[2]{*}{Noise Level} & \multicolumn{6}{c}{Hyperparameter} \bigstrut[t]\\
      &       & $T$   & $N^\prime$ & $\alpha$ & $\beta$ & $\gamma$ & $\lambda$ \bigstrut[b]\\
\hline
\hline
\multirow{4}[2]{*}{VOC} & 10\% & 0.05  & 10    & 10    & 0.7   & 0.3   & 0.05 \bigstrut[t]\\
      & 20\% & 0.05  & 10    & 10    & 0.7   & 0.3   & 0.05 \\
      & 30\% & 0.1   & 10    & 10    & 0.8   & 0.3   & 0.1 \\
      & 40\% & 0.1   & 10    & 5     & 0.8   & 0.3   & 0.1 \bigstrut[b]\\
\hline
\multirow{2}[2]{*}{COCO} & 20\% & 0.01  & 10    & 10    & 0.7   & 0.3   & 0.01 \bigstrut[t]\\
      & 40\% & 0.1   & 10    & 5     & 0.8   & 0.3   & 0.1 \bigstrut[b]\\
\specialrule{0.15em}{0pt}{0pt}
\end{tabular}%

}}    
     \caption{\textbf{Hyperparameter selections.}
     We report the hyperparameters for all settings to ensure reproducibility.
     }
    \label{tab:selection}
\end{table}


\section{More Ablation Studies}
\label{app:more_abl}
In this section, we conduct more ablation studies to further verify the effectiveness of the proposed DISCO.
These ablation studies contain hyperparameter sensitivity, backbone compatibility, and the execution number of DISCO.
Unless otherwise specified, the following experiments are all based on VOC at the $40\%$ noise level.

\begin{table}[t]
  \centering
  \setlength{\tabcolsep}{1pt}
  \renewcommand{\arraystretch}{1.6}

\begin{minipage}[c]{0.2\linewidth}
    \centering
    \resizebox{\linewidth}{!}{\setlength{\tabcolsep}{2mm}{
  
\begin{tabular}{c|c|c}
\specialrule{0.15em}{0pt}{0pt}
Hyper. & Value & $\text{AP}_{50}$ \bigstrut\\
\hline
\hline
\multirow{3}[2]{*}{$T$} & 0.01  &  68.6 \\
      & 0.1   & \textbf{68.7} \\
      & 0.2     &  67.8 \\
\specialrule{0.15em}{0pt}{0pt}
\end{tabular}
    }}
\end{minipage}
\begin{minipage}[c]{0.19\linewidth}
    \centering
    \resizebox{\linewidth}{!}{\setlength{\tabcolsep}{2mm}{
\begin{tabular}{c|c|c}
\specialrule{0.15em}{0pt}{0pt}
Hyper. & Value & $\text{AP}_{50}$ \bigstrut\\
\hline
\hline
\multirow{3}[2]{*}{$N^\prime$} & 5     & 68.5 \bigstrut[t]\\
      & 10    & \textbf{68.7} \\
      & 20    & 68.6 \bigstrut[b]\\
\specialrule{0.15em}{0pt}{0pt}
\end{tabular}
    }}
\end{minipage}
\begin{minipage}[c]{0.2\linewidth}
    \centering
    \resizebox{1\linewidth}{!}{\setlength{\tabcolsep}{2mm}{

\begin{tabular}{c|c|c}
\specialrule{0.15em}{0pt}{0pt}
Hyper. & Value & $\text{AP}_{50}$ \bigstrut\\
\hline
\hline
\multirow{3}[2]{*}{$\alpha$} & 3     &  68.3 \\
      & 5     & \textbf{68.7} \\
      & 7     &  68.2 \\
\specialrule{0.15em}{0pt}{0pt}
\end{tabular}
    }}
\end{minipage}

\begin{minipage}[c]{0.2\linewidth}
    \centering
    \resizebox{\linewidth}{!}{\setlength{\tabcolsep}{2mm}{

\begin{tabular}{c|c|c}
\specialrule{0.15em}{0pt}{0pt}
Hyper. & Value & $\text{AP}_{50}$ \bigstrut\\
\hline
\hline
\multirow{3}[2]{*}{$\beta$} & 0.7   &  68.1 \\
      & 0.8   & \textbf{68.7} \\
      & 0.9   &  67.3 \\
\specialrule{0.15em}{0pt}{0pt}
\end{tabular}%
    }}
\end{minipage}
\begin{minipage}[c]{0.19\linewidth}
    \centering
    \resizebox{1\linewidth}{!}{\setlength{\tabcolsep}{2mm}{
\begin{tabular}{c|c|c}
\specialrule{0.15em}{0pt}{0pt}
Hyper. & Value & $\text{AP}_{50}$ \bigstrut\\
\hline
\hline
\multirow{3}[2]{*}{$\gamma$} & 0.1   & 68.3 \bigstrut[t]\\
      & 0.3   & \textbf{68.7} \\
      & 0.5   & 68.4 \bigstrut[b]\\
\specialrule{0.15em}{0pt}{0pt}
\end{tabular}%
    }}
\end{minipage}
\begin{minipage}[c]{0.2\linewidth}
    \centering
    \resizebox{\linewidth}{!}{\setlength{\tabcolsep}{2mm}{
\begin{tabular}{c|c|c}
\specialrule{0.15em}{0pt}{0pt}
Hyper. & Value & $\text{AP}_{50}$ \bigstrut\\
\hline
\hline
\multirow{3}[2]{*}{$\lambda$} & 0.05  & 67.9 \bigstrut[t]\\
      & 0.1   & \textbf{68.7} \\
      & 0.15  & 67.4 \bigstrut[b]\\
\specialrule{0.15em}{0pt}{0pt}
\end{tabular}
    }}
\end{minipage}

     \caption{\textbf{Ablation studies of hyperparameter sensitivity.}
     DISCO can still achieve relatively stable performance when these hyperparameters vary within a moderate range.
     }
     \vspace{-3mm}
    \label{tab:sensitivity}
\end{table}

\subsection{Hyperparameter Sensitivity}

Here we evaluate the sensitivity of the hyperparameters used in DISCO.
Note that we choose some moderate values rather than extreme ones to reasonably evaluate the sensitivity of each hyperparameter.
The experimental results are reported in Table~\ref{tab:sensitivity}.
As we can see, the temperature coefficient $T$ is relatively robust when set to 0.01 or 0.2.
Tuning $T$ to a proper value can contribute to better performance.
Besides, it can be observed that the augmented proposal number $N^\prime$ is insensitive when varying from $5$ to $20$.
This is the reason why we empirically fix $N^\prime$ to $10$ for all settings.
Moreover, the hyperparameters regulate the fusion of two bounding boxes (\textit{i.e.}, $\alpha$ and $\beta$) is also insensitive when varying within a moderate range, showing the effectiveness of our method.
Furthermore, two loss weights $\gamma, \lambda$ also remain insensitive while $\lambda$ is relatively crucial.
This is because it controls the strength of an extra classification loss term, directly affecting classification accuracy.

\subsection{Backbone Compatibility}
Following \cite{liu2022robust}, the benchmark experiments are performed with ResNet-50 \cite{he2016deep} as the backbone.
To further demonstrate the superior performance of our method, we conduct an additional experiment based on different backbones.
Specifically, in this experiment, DISCO \cite{liu2022robust} is compared to OA-MIL on COCO at the $40\%$ noise level with the backbone set to ResNet-101 \cite{he2016deep}, and other experiment setups remain the same.
In this way, we aim to evaluate the performance of our DISCO for a large-scale dataset when it is equipped with an advanced backbone.
The experimental results are reported in Table~\ref{tab:backbone}.
It can be observed that DISCO can further improve performance and still achieve SOTA results.

\begin{table}[t]
  \centering
  \setlength{\tabcolsep}{1pt}
  \renewcommand{\arraystretch}{1.6}
  \begin{minipage}[c]{0.65\linewidth}
    \resizebox{\linewidth}{!}{
    \setlength{\tabcolsep}{2mm}{
    \begin{tabular}{l|l|cccccc}
    \hline
    Method & Backbone & $\text{AP}$ & $\text{AP}_{50}$ & $\text{AP}_{75}$ & $\text{AP}_\text{S}$ & $\text{AP}_\text{M}$ & $\text{AP}_\text{L}$ \bigstrut\\
    \hline
    \hline
    OA-MIL \cite{liu2022robust} & \multirow{2}[2]{*}{ResNet-50} & 18.6  & 42.6  & 12.9  & 9.2   & 19.0  & 26.5 \bigstrut[t]\\
    DISCO (Ours) &       & \textbf{21.2} & \textbf{45.7} & \textbf{16.9} & \textbf{11.4} & \textbf{24.7} & \textbf{27.8} \bigstrut[b]\\
    \hline
    OA-MIL \cite{liu2022robust} & \multirow{2}[2]{*}{ResNet-101} & 19.3  & 44.1  & 13.1  & 9.3   & 20.8  & 27.8 \bigstrut[t]\\
    DISCO (Ours) &       & \textbf{22.7} & \textbf{47.6} & \textbf{18.4} & \textbf{12.9} & \textbf{26.6} & \textbf{29.8} \bigstrut[b]\\
    \hline
    \end{tabular}%
    }}
    \end{minipage}
    \begin{minipage}[c]{0.29\linewidth}
      \resizebox{1\linewidth}{!}{
    \setlength{\tabcolsep}{4mm}{
    \begin{tabular}{c|c}
    \specialrule{0.15em}{0pt}{0pt}
    Execution Num. & $\text{AP}_{50}$ \bigstrut\\
    \hline
    \hline
    1     & 68.1 \bigstrut[t]\\
    2     & \textbf{68.7} \\
    3     & 67.9 \bigstrut[b]\\
    \specialrule{0.15em}{0pt}{0pt}
    \end{tabular}
    }}   
    \end{minipage}
     \caption{Left: \textbf{Ablation studies of backbone compatibility.}
     The experiment is conducted on COCO at 40\% noise level with ResNet-50 and ResNet-101.
     DISCO can still outperform OA-MIL when equipped with different backbones.
     Right: \textbf{Ablation studies of the execution number of DISCO.}
     Our execution strategy can achieve superior performance.
     }
    \vspace{-3mm}
    \label{tab:backbone}
\end{table}

\subsection{Execution Number of DISCO}
In this work, DISCO is performed twice in a training iteration, where the first time is for proposal re-assignment and the second time is for obtaining better supervision.
We compare such an execution strategy with two other options:
1) The execution number of DISCO is set to $1$: proposal re-assignment is removed and the only one time of DISCO is for obtaining better supervision;
2) The execution number of DISCO is set to $3$: the first two times are for proposal re-assignment and the third time is for obtaining better supervision.
As shown in Table~\ref{tab:backbone}, more execution numbers of DISCO do not contribute to better detection performance.
This is because such an improper strategy could result in excessive box refinement and thus influence the learning stability of detectors.
Moreover, it also can be observed that our execution strategy can achieve superior performance.


\section{More Qualitative Results}

\subsection{Box Refinement}
\begin{figure*}[t]
    \centering
    \includegraphics[width=1\linewidth]{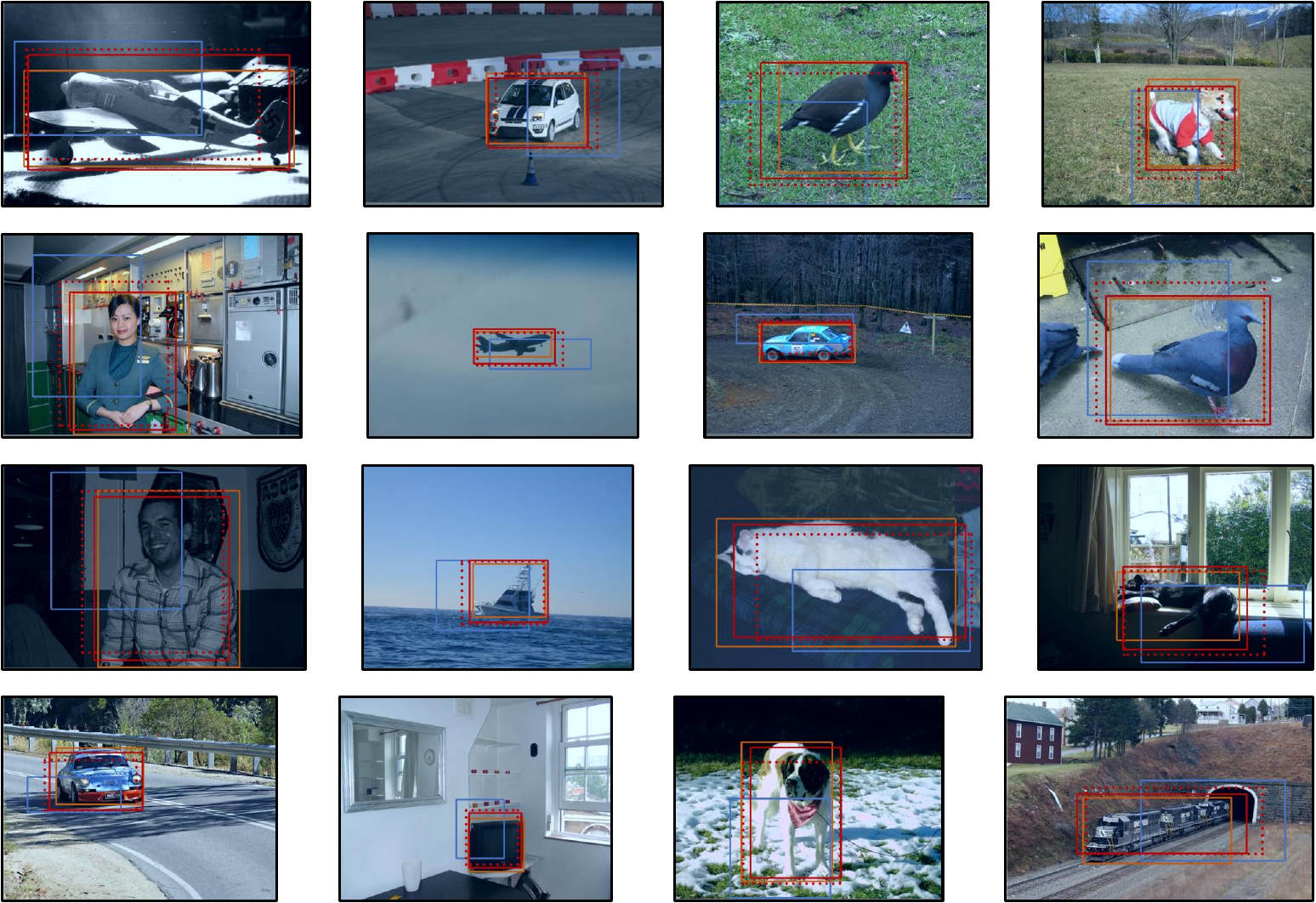}
    \caption{\textbf{Qualitative results of box refinement in DISCO.}
        Real ground-truths and noisy ground-truths are marked in \textit{orange} and \textit{blue}.
        Refined bounding boxes produced by the first-/second-time DISCO are indicated in \textit{dotted/solid red}.
        The first-time refined boxes can cover the objects more tightly than noisy ground-truths, and the second-time refinement can further contribute to more precise ones.
    }
    \label{fig:localization_more}
    \vspace{-2mm}
\end{figure*}

As an extension to Figure~\ref{fig:localization}, we present more qualitative results of box refinement in DISCO (see Figure~\ref{fig:localization_more}), which shows that DISCO can attain tighter bounding boxes than noisy ground-truths.
As shown in Figure~\ref{fig:localization_more}, it is worth noting that DISCO can achieve consistent refinement of bounding boxes for different objects varying in size.

\subsection{Interpretability}
\begin{figure*}[t]
    \centering
    \includegraphics[width=1\linewidth]{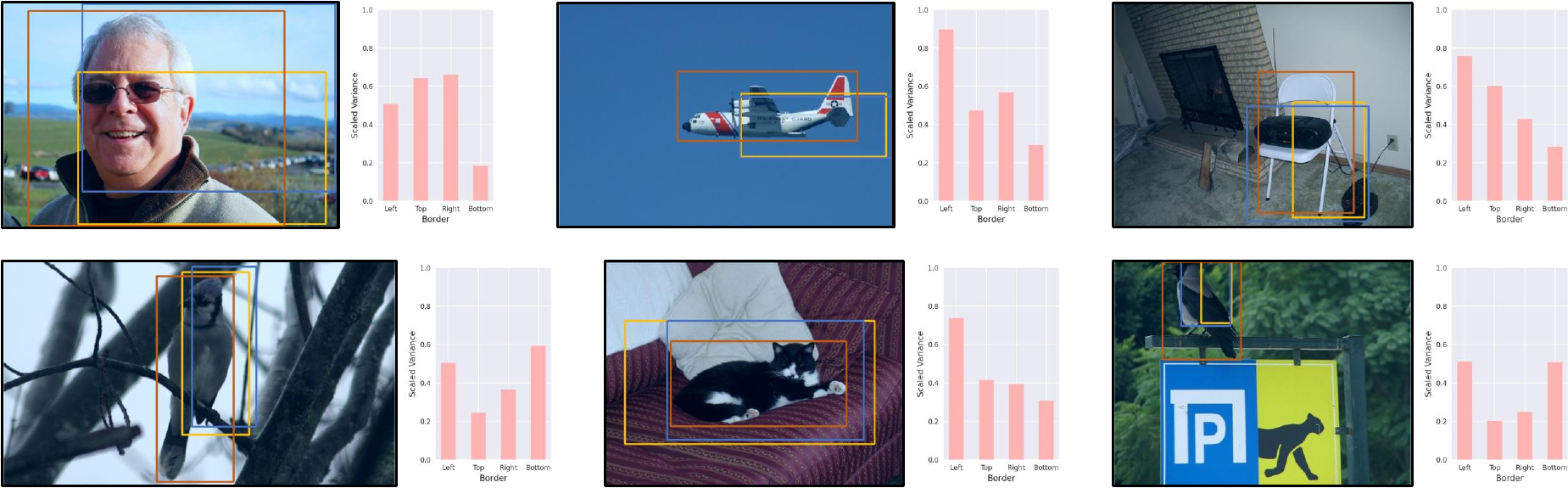}
    \caption{\textbf{Qualitative results of interpretability in DISCO.}
        We randomly choose an assigned proposal (\textit{yellow}) per image to report its estimated variances.
        Real ground-truths and noisy ground-truths are marked in \textit{orange} and \textit{blue}.
        Note that the variance is scaled by the width and height for clarity.
        With the proposed DA-Est, DISCO can estimate reasonable variances for each border of box prediction.
    }
    \label{fig:interpretability_more}
\end{figure*}

In Figure~\ref{fig:interpretability_more}, more qualitative results of interpretability in DISCO are provided to demonstrate such a characteristic of our method.
As shown in Figure~\ref{fig:interpretability_more},  when trained with DISCO, the detector can output a reasonable variance as the confidence for each border of predicted bounding boxes, which shows that the detector is capable of realizing which border may be inaccurately predicted.

\end{document}